# An Application of the Generalized Rectangular Fuzzy Model to Critical Thinking Assessment

Igor Ya. Subbotin\*, Michael Gr. Voskoglou\*\*

## Abstract

The authors apply the Generalized Rectangular Model (GRM) for assessing students' critical thinking skills. GRM is a variation of the center of gravity (COG) defuzzification technique, which was properly adapted and used by them several times in the past as an assessment method, called here the Rectangular Model (RM). The central idea of the GRM is the "movement" to the left of the rectangles appearing in the membership function's graph of the RM, thus making the adjacent rectangles to share common parts. This treatment reflects better than RM the ambiguous assessment cases of student scores being at the boundaries between two successive assessment grades (e.g. something like 84-85% being at the boundaries between A and B) and therefore belonging to the common parts of the above rectangles. In fact, in GRM, and in contrast to RM, assuming that these scores belong to both of the successive assessment grades, we consider twice the common parts of the rectangles for calculating the COG of the level's section lying between the resulting graph and the OX axis. Our results are illustrated on the data of a classroom application performed in one of the Los Angeles Unified District High Schools and connecting the students CT skills with their language competencies.

## 1. Introduction

In our modern society the composite problems of the day to day life require a higher-order thinking for their solution, which can be conceptualised as a complex mode of thinking that often generates multiple solutions. Such kind of thinking, usually referred as *Critical Thinking* (CT) involves synthesis and analysis, abstraction, uncertainty, application of multiple criteria, reflection, decision making, drawing warrant conclusions and generalizations, self-regulation, etc. It also facilitates the transfer of knowledge, i.e. the use and transformation of already existing knowledge for creating new knowledge.

The complexity of CT is evident from the fact that there is no definition that is universally accepted. Some of the most characteristic definitions for CT existing in the literature are the following: "…disciplined thinking that is clear, rational, open-minded, and informed by evidence" [1]; "the skill and propensity to engage in an activity with reflective skepticism" [3]; "…disciplined, self-directed thinking which exemplifies the perfection of thinking appropriate to a particular mode of domain of thinking." [4], etc.

In an earlier paper [8] we have used a properly adapted version of the *Center of Gravity (COG) defuzzification technique* for assessing student CT skills. Here, we call this approach the *Rectangular Model* (RM), since the membership function of the corresponding fuzzy set has a graph consisting of the sum of five rectangles (see Figure 1 below). Our target in the present paper is to assess the student CT skills by applying a recently developed [12] generalization of the RM which fits better to the ambiguous assessment cases.; we have called it the *Generalized Rectangular Model* (GRM)

The rest of the paper is formulated as follows: In Section 2 we give a brief account of our previous researches on the use of principles of *Fuzzy Logic* (FL) for student assessment. In Section 3 we present the GRM , while in Section 4 we compare it with the assessment methods developed and/or utilized in earlier authors' works and we make some important generalizations. In Section 5 we apply the GRM and the earlier used by the authors assessment methods on the data of a classroom application performed in one of the Los Angeles Unified District High Schools and connecting students' CT skills with their language competencies. Finally, Section 6 is devoted to our conclusions and to a brief discussion on our future plans for further research on the subject.

For general facts on *Fuzzy Sets* (FS) we refer to the book [2].

## 2. Summary of our previous researches

FL, due to its inherent property of characterizing the ambiguous situations with multiple values, offers a rich source of assessment perspectives. In 1999 Voskoglou [14] developed a fuzzy model for the description of the learning process by representing its main steps as FSs on the set $U$ = {A, B, C, D, F} of linguistic labels (grades) characterizing the individuals' learning performance, where A (85-100%) = excellent, B (75-84%) = very good, C(60-74%) = good, D(50-59%) = fair and F(<50%) = unsatisfactory [1]. Also, in a later work [15] Voskoglou used the corresponding fuzzy system's *total uncertainty* as a measure for assessing the student performance in learning mathematics. Meanwhile Subbotin et al. [5], based on Voskoglou's [14] model, adopted properly the well known in fuzzy mathematics COG defuzzification technique [13] to provide an alternative assessment measure of student learning skills, thus creating the RM.

More explicitly, the process of *reasoning with fuzzy rules* involves:
- *Fuzzification* of the problem's data by utilizing the suitable membership functions to define the required FSs.
- *Application of FL operators* on the defined FSs and combination of them to obtain the final result in the form of a unique FS.
- *Defuzzification* of the final FS to return to a crisp output value, in order to apply it on the real world situation for resolving the corresponding problem.

There are several methods reported in the literature for performing the process of defuzzification. For example, the calculation of the fuzzy system's total uncertainty, applied by Voskoglou in [15] for measuring learning skills, could actually be considered as a deffuzification method. On the other hand, the COG technique, the most popular such method in fuzzy mathematics, *r*eplaces the elements of the universal set $U$ by prefixed real intervals in order to enable the construction of the graph of the corresponding membership function and uses the coordinates of the COG of the level's section defined by this graph and the OX axis in order to obtain the required crisp output value (e.g. see [13], etc).

---

[1] The scores corresponding to the linguistic grades are not standard; they may differ from case to case according to the assessor's personal criteria. For example, in a more strict assessment we could have A (90-100%), B (80-89%), C(65-79%), D(55-64%), F(<55%), etc.

Subbotin et al. in 2004 [5] represented the group G of students under assessment as a FS on the set $U$ = {A, B, C, D, F} of the Voskoglou's model for learning [14] in the form G = {(x, m(x)): $x \in U$}, where y=m(x): $U \to$ [0, 1] is the corresponding membership function, and replaced the linguistic labels of $U$ by real intervals as follows: F $\to$ [0, 1), D $\to$ [1, 2), C $\to$ [2, 3), B $\to$ [3, 4), A $\to$ [4, 5] . Then, we have $y_1 = m(x) = m(F)$ for all $x$ in [0,1), $y_2 = m(x) = m(D)$ for all $x$ in [1,2), $y_3 = m(x) = m(C)$ for all $x$ in [2, 3), $y_4 = m(x) = m(B)$ for all $x$ in [3, 4) and $y_5 = m(x) = m(A)$ for all $x$ in [4,5). Therefore, the graph of the membership function $y = m(x)$ takes the form of the bar graph of Figure 1. The area $S$ of the level's section defined by this graph and the OX axis is equal to the sum of the areas $S_i$, i =1, 2, 3, 4, 5, of five *rectangles,* whose sides lying on the OX axis have length equal to 1 metric unit .
.

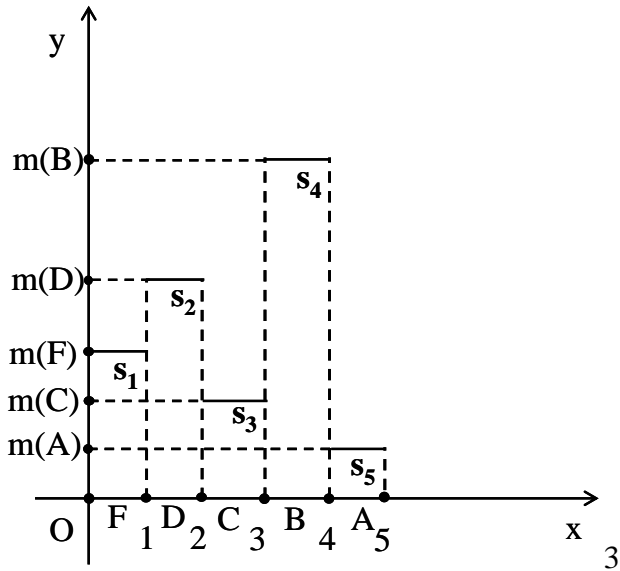

**Figure 1:** Bar graphical data representation of the RM

Then, using the well known from Mechanics formulas

$$X_c = \frac{\iint_S x\,dxdy}{\iint_S dxdy}, Y_c = \frac{\iint_S y\,dxdy}{\iint_S dxdy} \quad (1)$$

it is straightforward to check that the coordinates ($X_c$, $Y_c$) of the COG of the area $S$ are calculated by the formulas :

$$X_c = \frac{1}{2}(y_1+3y_2+5y_3+7y_4+9y_5), \quad Y_c = \frac{1}{2}(y_1^2+y_2^2+y_3^2+y_4^2+y_5^2) \quad (2),$$

with $y_i = \dfrac{m(x_i)}{\sum_{j=1}^{5} m(x_j)}$, where $x_1$=F, $x_2$=D, $x_3$=C, $x_4$=B and $x_5$=A.

Further, applying elementary algebraic inequalities one can determine the area (a triangle) where the COG lies and by elementary geometric observations on the corresponding triangle can obtain a criterion for comparing the performances of different student groups (see [5] or Section 4 of [19]). A similar argument will be applied in Section 3 below for the development of the GRM.

Since then (i.e. after 2004) both authors of the present paper, either collaborating or independently to each other, utilized the RM for assessing other student competencies (e.g. see [8, 16, 17, 19], etc), for testing the effectiveness of a CBR system [6]., for assessing the Bridge players' performance ([18] and Section 6.2 of [19]), etc.

Recently, Subbotin & Bilotskii [7] introduced a *Triangular Fuzzy Assessment Model* (TFAM) for assessing the students' learning skills, which was fully developed by the present authors in [9]. The basic idea of the TFAM is the replacement of the rectangles appearing in the membership function's graph of the RM (Figure 1) by isosceles triangles sharing common parts (see Figure 4 of [19]). In this way one treats better the ambiguous cases of student scores being at the boundaries between two successive linguistic grades (e.g. something like 84-85% being at the boundaries between A and B). An alternative version of the RM is the *Trapezoidal Fuzzy Assessment Model* (TpFAM) initiated by Subbotin [10] and fully developed by the present authors in [11], in which the rectangles of the RM are replaced by isosceles trapezoids sharing common parts (see Figure 2 of [19]). The formulas calculating the coordinates $(X_c, Y_c)$ of the COG are:

$$\text{TFAM:} \quad X_c = (0.7\sum_{i=1}^{5} iy_i) - 0.2, \quad Y_c = \frac{1}{5}\sum_{i=1}^{5} y_i^2 \quad (3)$$

$$\text{TpFAM:} \quad X_c = (0.7\sum_{i=1}^{5} iy_i) - 0.2, \quad Y_c = \frac{3}{7}\sum_{i=1}^{5} y_i^2 \quad (4).$$

In both cases we have $x_1$ = F, $x_2$ = D, $x_3$ = C, $x_4$ = B and $x_5$ = A, while the values of the $y_i$'s are equal to the ratios (*frequencies*) of the numbers of students obtained the grade $x_i$, i = 1, 2, 3, 4, 5, to the total number of students under assessment.

All the above fuzzy assessment models (measurement of the Uncertainty, RM, TFAM and TpFAM) are presented in detail in [19] (see also Remark (ii) of Section 4), together with two applications (student and Bridge players' assessment), in which these models are validated by comparing their outcomes with the corresponding outcomes of two traditional assessment methods (calculation of the *mean values* and the *GPA index*) based on the principles of classical (bi-valued) logic. Keeping the above notation for the $x_i$'s and the $y_i$'s we recall that the *Grade Point Average* (GPA) index is calculated by the formula: GPA= $y_2 + 2y_3 + 3y_4 + 4y_5$ (5).

## 3. The Generalized Rectangular Model (GRM).

As it has been previously mentioned, the TFAM and TpFAM have been developed for reflecting the frequently appearing ambiguous assessment cases. The question however is: Is it necessary for this reason to change the shape of the assessment model (triangles or trapezoids instead of rectangles)? The effort of answering this question led to the development of the GRM [12]. The central idea of GRM is the 'movement" of the rectangles of the RM to the left, thus making the adjacent rectangles to share common parts (Figure 2).

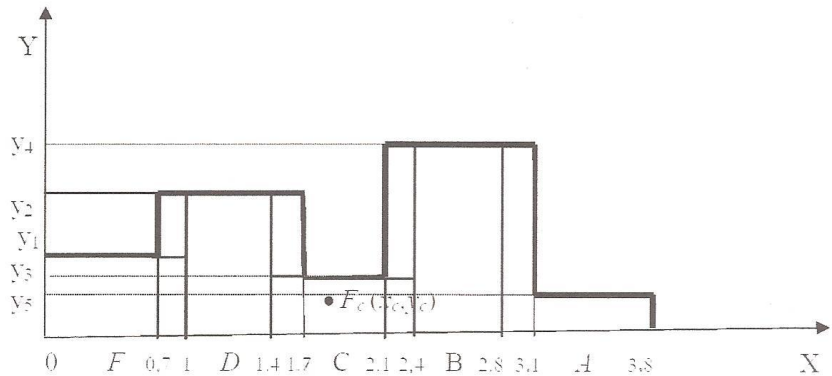

**Figure 2:** Bar graphical data representation of the GRM

The above treatment reflects better than RM the student scores being at the boundaries between two successive assessment grades and therefore belonging to the common parts of the adjacent rectangles. In fact, assuming (in contrast to RM) that these scores belong to both of the successive assessment grades, we *consider twice* the common parts of the rectangles for calculating the COG of the area lying between the resulting graph and the OX axis. Therefore, the COG here will be calculated not by applying formulas (1), as we did in the case of RM., but as the *resultant of the system of COGs of the five rectangles*. In detail, the whole process involves the following steps:

1. Let $y_1$, $y_2$, $y_3$, $y_4$, $y_5$ be the *frequencies* of students in the group who obtained the grades F, D, C, B, A respectively. Then $\sum_{i=1}^{5} y_i = 1$ (100%).

2. We take the heights of the rectangles in Figure 2 to have lengths equal to the corresponding student frequencies. Also, the sides of the adjacent rectangles lying on the OX axis have common parts with length equal to the 30% of their lengths, i.e. 0.3 units.

3. We calculate the coordinates ($x_{c_i}$, $y_{c_i}$) of the COG, say $F_i$, of each rectangle, i=1, 2, 3, 4, 5 as follows: Since the COG of a rectangle is the point of the intersection of its diagonals, we have that $y_{c_i} = \frac{1}{2} y_i$. Also, since the x-coordinate of each $F_i$ is equal to the

x- coordinate of the middle of the side of the corresponding rectangular lying on the OX axis, from Figure 2 it is easy to observe that $x_{c_i} = 0.7i - 0.2$.

4. We consider the system of the COGs $F_i$ and we calculate the coordinates $(X_c, Y_c)$ of the COG F of the whole area considered in Figure 2 as the resultant of the system of the GOCs $F_i$ of the five rectangles from the following well known formulas [20]:
$X_c = \frac{1}{S}\sum_{i=1}^{5} S_i x_{c_i}$, $Y_c = \frac{1}{S}\sum_{i=1}^{5} S_i y_{c_i}$ (6). In the above formulas $S_i$, i= 1, 2, 3, 4, 5 denote the areas of the corresponding rectangles, which are equal to $y_i$. Therefore $S = \sum_{i=1}^{5} S_i = \sum_{i=1}^{5} y_i = 1$ and formulas (6) give that $X_c = \sum_{i=1}^{5} y_i(0.7i - 0.2)$, $Y_c = \sum_{i=1}^{5} y_i(\frac{1}{2} y_i)$ or

$$X_c = (0.7\sum_{i=1}^{5} i y_i) - 0.2, \quad Y_c = \frac{1}{2}\sum_{i=1}^{5} y_i^2 \quad (7).$$

5. We determine the area where the COG F lies as follows: For i, j=1, 2, 3, 4, 5, we have that $0 \leq (y_i - y_j)^2 = y_i^2 + y_j^2 - 2y_i y_j$, therefore $y_i^2 + y_j^2 \geq 2 y_i y_j$, with the equality holding if, and only if, $y_i = y_j$. Therefore $1 = (\sum_{i=1}^{5} y_i)^2 = \sum_{i=1}^{5} y_i^2 + 2\sum_{\substack{i,j=1,\\i\neq j}}^{5} y_i y_j \leq \sum_{i=1}^{5} y_i^2 + 2\sum_{\substack{i,j=1,\\i\neq j}}^{5} (y_i^2 + y_j^2)$

$= 5\sum_{i=1}^{5} y_i^2$ or $\sum_{i=1}^{5} y_i^2 \geq \frac{1}{5}$ (8), with the equality holding if, and only if, $y_1 = y_2 = y_3 = y_4 = y_5 = \frac{1}{5}$. In the case of the equality the first of formulas (7) gives that $X_c = 0.7(\frac{1}{5} + \frac{2}{5} + \frac{3}{5} + \frac{4}{5} + \frac{5}{5}) - 2 = 1.9$. Further, combining the inequality (8) with the second of formulas (7) one finds that $Y_c \geq \frac{1}{10}$ Therefore the *unique minimum for $Y_c$* corresponds to the COG $F_m$(1.9, 0.1). The *ideal case* is when $y_1 = y_2 = y_3 = y_4 = 0$ and $y_5 = 1$. Then formulas (7) give that $X_c = 3.3$ and $Y_c = \frac{1}{2}$. Therefore the COG in this case is the point $F_i$(3.3, 0.5). On the other hand, the *worst case* is when $y_1 = 1$ and $y_2 = y_3 = y_4 = y_5 = 0$. Then from formulas (7) we find that the COG is the point $F_w$(0.5, 0.5). Therefore, *the area where the COG F lies is the area of the triangle* $F_w$ $F_m$ $F_i$ (Figure 3). Then from elementary geometric observations it follows that the greater is the value of $X_c$, the better is the group's performance. Also, for two groups with the same $X_c \geq 1.9$, the group having the COG which is situated closer to $F_i$ is the group with the higher $Y_c$ and for two groups with the same $X_c < 1.9$ the group having the COG which is situated farther to $F_w$ is the group with the lower $Y_c$. Based on the above considerations we formulate our criterion for comparing the groups' performances in the following form:

- *Between two groups, the group with the greater $X_c$ performs better.*
- *If two groups have the same $X_c \geq 1.9$, then the group with the greater $Y_c$ performs better.*

- *If two groups have the same $X_c < 1.9$, then the group with the lower $Y_c$ performs better.*

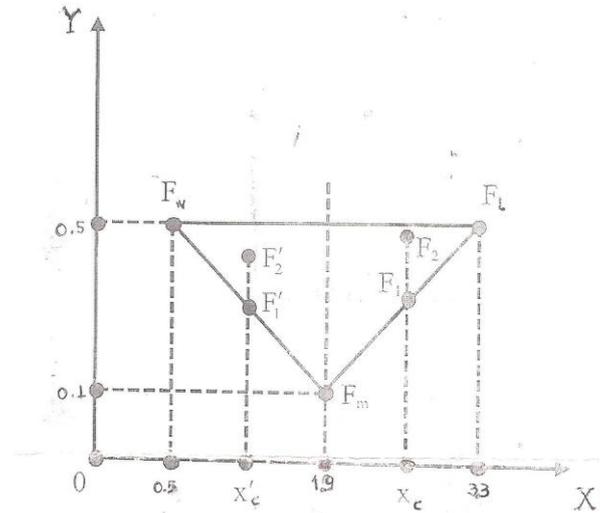

**Figure 3:** The triangle where the COG lies

## 4. Remarks and generalizations

   **(i)** In case of the RM the membership function $y = m(x)$ in formulas (2) could be defined in any, compatible to the common logic, way [2]. However, in order to be able to compare the assessment conclusions obtained by all the above mentioned assessment methods, we define here $y = m(x)$ in terms of the student frequencies, as we did for GPA, RM, TFAM, TpFAM and GRM (see Sections 2 and 3).

   **(ii)** Combining formulas (2), (3), (4), (5) and (7) with the corresponding assessment criteria, it becomes evident that the GPA index and the RM, TFAM, TpFAM, GRM assess the *quality performance* of the student groups' by assigning greater coefficients (weights) to the higher grades. On the contrary, the corresponding fuzzy system's uncertainty and the traditional calculation of the mean values assess the groups' *mean performance*.

   **(iii)** In Section 5 of [19] we assigned to the grade F the interval [0, 10] and letting the bases of the adjacent figures on the OX-axis to share 30% of their lengths we considered the graphs for the construction of the TpFAM and the TFAM on the interval [0, 38]. In fact, the interval [7, 17] was assigned to D, the interval [14, 24] to C, the interval [21, 31] to B and the interval [28, 38] to A. On the contrary, in the present paper we have assigned to F the interval [0, 1] and therefore we have considered the graphs for the construction of

---

[2] We recall that, for defining a FS, the choice of the membership function *is not unique*, depending on the constructor's subjective criteria . However, a necessary condition for the creditability of the corresponding fuzzy model in representing the real situation is that the membership function's definition must be compatible to the common logic.

the TFAM, TpFAM and GRM on the interval [0, 3.8] (see Figure 2). As a result of these manipulations, in [19] the first of formulas (3) and (4) were proved to be

$X_c = (7\sum_{i=1}^{5} iy_i) - 2$, instead of the form $X_c = (0.7\sum_{i=1}^{5} iy_i) - 0.2$, which is given here.

**(iv)** We can write formulas (3), (4) and (7) in the *single form*:

$$X_c = (0.7\sum_{i=1}^{5} iy_i) - 0.2 \, , \, Y_c = a\sum_{i=1}^{5} y_i^2 \quad (8),$$

with $a = \frac{1}{5}$ for the TFAM, $a = \frac{3}{7}$ for the TpFAM and $a = \frac{1}{2}$ for the GRM. Observe that in all these formulas we deal with the same *key expressions* $\sum_{i=1}^{5} iy_i$ for $X_c$ and $\sum_{i=1}^{5} y_i^2$ for $Y_c$.

**(v)** It is easy to check that the above mentioned key expressions remain the same for percentages k% of the common lengths, with $k \neq 30$. Note that, since the ambiguous assessment cases are situated at the boundaries of the adjacent grades, which means that their x-coordinates take values near to the end of the corresponding to these grades real intervals, it is logical to accept that k < 50. Moreover, if $k \geq 50$, then in Figure 2 the interval of C on the OX axis will be completely covered by the intervals of D and B.

In general, assigning the interval [0, 1] to F and considering our graphs on the interval [0, m] and the bases of the adjacent figures sharing *k*% of their lengths, it is easy to check that m = 5 - $\frac{4k}{100}$. In fact, initially the interval [0, 5] is needed to construct the five rectangles of the RM (or triangles of the TFAM, or trapezoids of the TpFAM), while the movement of these figures to the left for sharing common parts reduces the length (5 units) of this interval by $\frac{4k}{100}$ units [3].

Consequently, for the comparing purposes one can reform the assessment criteria obtained for the TFAM/TpFAM (e.g. see Section 5 of [19]) and for the GRM (see Section 3) in the following *unified form* based on the above mentioned key expressions:

- *Between two groups the group with the greater value of $\sum_{i=1}^{5} iy_i$ ($X_c$) performs better.*

---

[3] If we assign to F the interval [0, 10], then a similar argument shows that m = 50 - $\frac{4k}{10}$. It is also possible to consider n assessment grades with $n \neq 5$; in this case we obviously have that m = n - $\frac{(n-1)k}{100}$.

- *If two groups have the same $X_c \geq 0.5$m, then the group with the greater value of $\sum_{i=1}^{5} y_i^2$ ($Y_c$) performs better.*
- *If two groups have the same $X_c < 0.5$m, then the group with the lower value of $\sum_{i=1}^{5} y_i^2$ ($Y_c$) performs better.*

The above unified criterion shows that the TFAM, the TpFAM and the GRM are *equivalent assessment models*, in the sense that they obtain the same conclusions. This means that it makes no difference which one is used; the choice depends on the user's personal criteria.

**(vi)** Combining formulas (2), (5) and (7) [or (3), or (4)] with the corresponding assessment criteria we form the following Table containing the coefficients (weights) assigned by the RM, the GPA index and the GRM (or the equivalent to it TFAM and TpFAM) to the higher grades A and B.

**Table 1:** Coefficients of the higher scores

| Assessment Model | A | B |
|---|---|---|
| RM | 3.5 | 4.5 |
| GPA | 3 | 4 |
| GRM | 2.8 | 3.5 |

A simple inspection of Table 1 shows that RM is the most sensitive and GRM is the less sensitive to the higher scores model. This in practice means that, although all these models assess the quality performance of the student groups, in certain cases *differences can appear* to their assessment conclusions.

**(vii)** We can write $\sum_{i=1}^{5} i y_i = \sum_{i=1}^{5} y_i + (y_2 + 2y_3 + 3y_4 + 4y_5) = 1 + $ GPA. This, combined with the assessment criterion in (iii), shows that for two student groups with *different GPA values* the assessment conclusions obtained by GRM, TFAM and TpFAM are the *same* with those obtained by the application of the GPA index. However, in case of the *same GPA values* the application of the GPA index *could not lead to logically based conclusions*. In such situations, our criterion in (iii), due to its logical nature, becomes useful. For illustrating this, let us consider the following simple, but characteristic example, concerning two Classes with 60 students in each Class

**Table 2:** Student grades

| Grades | Class I | Class II |
|--------|---------|----------|
| C      | 10      | 0        |
| B      | 0       | 20       |
| A      | 50      | 40       |

The GPA index for the two classes is $\frac{3*10+5*50}{60} = \frac{4*20+5*40}{60} \approx 4.67$, which means that the two Classes demonstrate the same quality performance. On the contrary, applying the criterion of (iii) one finds that $X_c \approx 0.7*4.67 - 0.2 = 3.069 > 1.9$ for both Classes, but $\sum_{i=1}^{5} y_i^2 = (\frac{1}{6})^2 + (\frac{5}{6})^2 = \frac{26}{36}$ for the first and similarly $\sum_{i=1}^{5} y_i^2 = \frac{20}{36}$ for the second Class. Therefore the Class I performed better.

Now which one of the above two conclusions is closer to the reality? For answering this question, let us consider first the *quality of knowledge*, i.e. the ratio of the students received B or better to the total number of students, which is equal to $\frac{5}{6}$ for the first and 1 for the second Class. Therefore, from the common point of view, the situation in Class II is better.

Also, if we assign to the grades A, B, C, D and F the commonly accepted numbers 5, 4, 3, 2 and 1 respectively, we find for Class I the mean values $\overline{X} = \frac{3*10+5*50}{60} \approx 4.67$ and $\overline{X^2} = \frac{3*10^2 + 5*50^2}{60} \approx 213.33$. Therefore the variance of $X$ is equal to $213.33 - (4.67)^2 \approx 191.52$. In the same way one finds that the variance of $X$ for Class II is equal to $160 - (4.67)^2 \approx 138.19 < 213.33$. Therefore the standard deviation for the second Class is definitely smaller, which means that, from the statistical point of view, the situation in Class II is also better.

However, some instructors could prefer the situation in Class I, which has much more "perfect" students. Everything is determined by the personal preference of the goals. The conclusion of the GRM agrees with the second point of view, while the conclusion of the GPA looks as not having any logical basis.

## 5. Assessing student CT skills: A classroom application

Beyond understanding theory and formulas, the students need to be proficient in application of mathematics and science knowledge to different situations and challenges. That is why just the well developed *reading comprehending skills* are crucially important for solving mathematical content problems. In fact, even being skillful in the formal technical mathematics and communication of mathematics, the student, whose reading comprehension abilities are limited, will not be able to make any progress in the application of these mathematical skills to some problems or just simple questions related to real world. There is no need to justify the above obvious statement. We just want to support it by the following interesting example:

In one of the Los Angeles Unified District inner city schools having very diverse student population (Hispanic 53% , Asian 22%, Black 18%, White 7%) the Algebra 2 District Assessment Test was given. The test contents can be found in the Appendix attached to the article. A very professional and dedicated teacher, who conducted this test, gave it in two of his Algebra 2 classes. The one of them was a *regular class*, while the other was a so-called *shelter class*, which means that the waist majority of the students in this class were students for whom English is a second language, not a native tongue. The results of the test are presented in Tables 1 and 2 below .for the shelter and the regular class respectively.

**Table 1**: Results of the shelter class

| % Scale | Grade | Students |
| --- | --- | --- |
| 85-100 | A | 0 |
| 75-84 | B | 5 |
| 60-74 | C | 6 |
| 50-59 | D | 9 |
| Less than 50 | F | 18 |
| Total | | 38 |

**Table 2:** Results of the regular class

| % Scale | Grade | Students |
| --- | --- | --- |
| 85-100 | A | 0 |
| 75-84 | B | 1 |
| 60-74 | C | 5 |
| 50-59 | D | 3 |
| Less than 50 | F | 20 |
| Total | | 29 |

Next, we shall apply the assessment approaches mentioned in Section 3, in order to compare the conclusions obtained in each case. We only omit the measurement of the uncertainty, which is based on complex formulas and needs laborious calculations (e.g. see [15]).

(i) *Mean values:* Assigning the values 5, 4, 3, 2, 1 to the grades A, B, C, D, F respectively one finds the means $\frac{5*4+6*3+9*2+18.1}{38} \approx 1.95$ for the shelter and $\frac{1*4+5*3+3*2+20*1}{29} \approx 1.55$ for the regular class respectively. Therefore, the shelter class demonstrated a better *mean performance*.

(ii) *GPA index:* Applying formula (5) one finds that GPA= $\frac{9*1+6*2+5*3}{38} \approx 0.95$ for the shelter and GPA= $\frac{3*1+5*2+1*3}{29} \approx 0.55$. From formula (5) one also obtains that in case of the *ideal performance* ($y_i = 0$ for i<5, $y_5 = 1$) GPA takes its maximal value 4. Here, since both values of the GPA index are smaller than 2 (the half of its maximal value), both classes demonstrated a less than satisfactory *quality performance*; however the shelter class demonstrated again a better performance.

(iii) *Application of the RM:* Defining the membership function $y = m(x)$ in terms of the student frequencies and applying the first of formulas (2) one finds that $X_c = \frac{1}{2} * \frac{18+3*9+5*6+7*5}{38} \approx 1.45$ and $X_c = \frac{1}{2} * \frac{20+3*3+5*5+7*1}{29} \approx 1.05$ for the shelter and the regular class respectively. Therefore, according to the corresponding criterion (see Section 4 of []) the shelter class demonstrates again a better performance. Further, since the value of $X_c$ for both classes is much smaller than the half of its value in the ideal case ($\frac{9}{2}: 2 = 2.25$), their quality performance is characterized as unsatisfactory.

(iv) *Application of the GRM (or of the equivalent to it TFAM and TpFAM):* The first of formulas (7) or (3) or (4) gives that

$X_c = (0.7 * \frac{18+2*9+3*6+4*5}{38}) - 0.2 \approx 1.16$ for the shelter and

$X_c = (0.7 * \frac{20+2*3+3*5+4*1}{29}) - 0.2 \approx 0,89$ for the regular class. Since both values of the $X_c$ are smaller than the half of its value in the ideal case (3.3: 2 = 1.65) the two classes demonstrated again a less than satisfactory quality performance, with the performance of the shelter class being better.

In concluding, it was logical to expect that the test's results of the shelter class would be worse than in the regular class. However, the application of all the above assessment methods shows that the situation was opposite. Surprisingly, the shelter class performed better. This happened because the teacher, taking into account that the students in this class were not proficient in English, worked constantly on a daily basis on developing the students' mathematics vocabulary and comprehension in reading mathematics content problems. This training affected student's critical thinking and problem solving abilities.

## 6. Discussion and conclusions

The methods of assessing a group's performance usually applied in practice are based on the principles of the bi-valued logic (yes-no). However, this approach is not the most suitable, when dealing with the frequently .appearing in practice ambiguous assessment situations. In such cases, FL, due to its nature of including multiple values, offers a wider and richer field of resources. This gave us the impulse to apply here the GRM, a generalized form of the RM (which is a variation of the COG defuzzification technique) reflecting better the ambiguous cases of student scores being at the boundaries between two successive assessment grades.

The conclusions of our classroom application, connecting the students' CT skills with their language competencies, provided a strong indication that the results obtained by applying of the GRM fit to the corresponding results of other assessment methods (traditional and fuzzy) developed and/or utilized in our older researches (validation of the GRM). However, there is a need for more classroom applications for obtaining safer statistical data. On the other hand, since the GRM approach has the potential of a general assessment method, our plans for future research include the effort of applying this approach for assessing several other human (e.g. learning, problem-solving, spiritual games and tests, sports, competitions, etc)  or machine (e.g. rule based and CBR systems, decision-making systems, etc) activities.

*Igor Ya. Subbotin, Ph.D., National University, Department of Mathematics and Natural Sciences, Los Angeles, California, USA*

***Michael Gr. Voskoglou, Ph.D., Graduate Technological Educational Institute (T. E. I.) of Western Greece, Department of Applied Mathematics, Patras, Greece*

**Appendix: Algebra 2 Periodic Assessment** [4]

*1. Sketch a graph to model each of the following four situations. Think about the shape of the graph and whether it should be a continuous line or not.*

**A**: Candle
Each hour a candle burns down the same amount. $x$ = the number of hours that have elapsed. $y$ = the height of the candle in inches.

**B**: Letter
When sending a letter, you pay quite a lot for letters, weighing up to an ounce. You then pay a smaller, fixed amount for each additional ounce (or part of an ounce.)
$x$ = the weight of the letter in ounces.
$y$ = the cost of sending the letter in cents.

**C**: Bus
A group of people rent a bus for a day. The total cost of the bus is shared equally among the passengers.
$x$ = the number of passengers.
$y$ = the cost for each passenger in dollars.

**D**: Car value
My car loses about half of its value each year.
$x$ = the time that has elapsed in years.
$y$ = the value of my car in dollars.

*2. The formulas below are models for the situations. Which situation goes with each formula? Write the correct letter (A, B, C or D) under each one.*

$Y = \dfrac{300}{x}$
$y = 12 - 0.5x.$
$y = 30 + 20x.$
$y = 2000 \cdot (0.5)^x.$

*3. Answer the following questions using the formulas. Under each answer show your reasoning.*

How long will the candle last before it burns completely away?
How much will it cost to send a letter weighing 8 ounces?
If 20 people go on the coach trip, how much will each have to pay?

---

[4] Retrieved from: *Student Materials Functions and Everyday Situations* © 2012 MARS, Shell Center, University of Nottingham..

How much will my car be worth after 2 years?